\documentclass[a4paper,fleqn]{cas-dc}

\usepackage{cleveref}
\usepackage[numbers]{natbib}
\usepackage{amssymb}
\usepackage{amsmath,amsfonts}
\usepackage{algorithmic}
\usepackage{algorithm}

\usepackage{array}
\usepackage{subcaption}
\usepackage{textcomp}
\usepackage{stfloats}
\usepackage{url}
\usepackage{verbatim}
\usepackage{graphicx}
\def\BibTeX{{\rm B\kern-.05em{\sc i\kern-.025em b}\kern-.08em
    T\kern-.1667em\lower.7ex\hbox{E}\kern-.125emX}}
\usepackage{color}
\usepackage{balance}
\usepackage{pifont}
\usepackage{tabularx}
\usepackage{booktabs}
\usepackage{multirow}
\usepackage{etoolbox}
\usepackage{bbm}
\usepackage{bm}
\usepackage{makecell}
\usepackage{dsfont} 

\makeatletter
\patchcmd{\@makecaption}
  {\scshape}
  {}
  {}
  {}
\makeatletter
\patchcmd{\@makecaption}
  {\\}
  {.\ }
  {}
  {}

\makeatother

\def\tsc#1{\csdef{#1}{\textsc{\lowercase{#1}}\xspace}}
\tsc{WGM}
\tsc{QE}

\usepackage{caption}
\hypersetup{
    colorlinks=true,
    citecolor=green,
    linkcolor=red,
    urlcolor=red
}

\begin{document}
\let\WriteBookmarks\relax
\def\floatpagepagefraction{1}
\def\textpagefraction{.001}
\let\printorcid\relax 
\shorttitle{OG-ReG}    
\shortauthors{Bohao Xing et al.}

\title[mode = title]{Insights from Visual Cognition: Understanding Human Action Dynamics with Overall Glance and Refined Gaze Transformer}  

\author[1]{Bohao Xing}[type=editor,
    auid=000]

\author[1]{Deng Li}
 
\author[1]{Rong Gao}

\author[1]{Xin Liu}
\ead{linuxsino@gmail.com} 
\cormark[1]

\author[1,2]{Heikki Kälviäinen}

\address[1]{Department of Computational Engineering, Lappeenranta-Lahti University of Technology LUT, Finland}
\address[2]{Faculty of Information Technology, Brno University of Technology, Czech Republic}

\cortext[1]{Corresponding author} 


\makeatletter\def\Hy@Warning#1{}\makeatother
\maketitle
\begin{abstract}
Recently, Transformer has made significant progress in various vision tasks. To balance computation and efficiency in video tasks, recent works heavily rely on factorized or window-based self-attention. However, these approaches split spatiotemporal correlations between regions of interest in videos, limiting the models' ability to capture motion and long-range dependencies. In this paper, we argue that, similar to the human visual system, the importance of temporal and spatial information varies across different time scales, and attention is allocated sparsely over time through glance and gaze behavior. Is equal consideration of time and space crucial for success in video tasks? Motivated by this understanding, we propose a dual-path network called the Overall Glance and Refined Gaze (OG-ReG) Transformer. The Glance path extracts coarse-grained overall spatiotemporal information, while the Gaze path supplements the Glance path by providing local details. Our model achieves state-of-the-art results on the Kinetics-400, Something-Something v2, and Diving-48, demonstrating its competitive performance. The code will be available at \href{https://github.com/linuxsino/OG-ReG}{https://github.com/linuxsino/OG-ReG}. 
\end{abstract}

\section{Introduction}
Action recognition is a crucial task that has garnered extensive attention in video understanding. Convolutional Neural Networks (CNNs) \cite{simonyan2014two, wang2016temporal, tran2015learning, ILSVRC15, carreira2017quo, qiu2017learning, tran2018closer, xie2018rethinking, lin2019tsm, feichtenhofer2019slowfast, feichtenhofer2020x3d, kwon2020motionsqueeze} have made remarkable progress in this field. More recently, the introduction of Transformer into computer vision \cite{dosovitskiy2020image} has led to great success in various image tasks \cite{touvron2021training, liu2021swin, strudel2021segmenter, carion2020end, zhu2020deformable, li2024enhancing, gao2026identity}. Subsequently, research on Transformer has been extended to video tasks \cite{bertasius2021space, neimark2021video, arnab2021vivit, li2026msf, liu2022video}. However, due to a large number of video tokens, vanilla self-attention becomes computationally expensive, making it impractical for extracting overall spatiotemporal features.


\begin{figure*}
  \centering
  \includegraphics[width=0.9\linewidth]{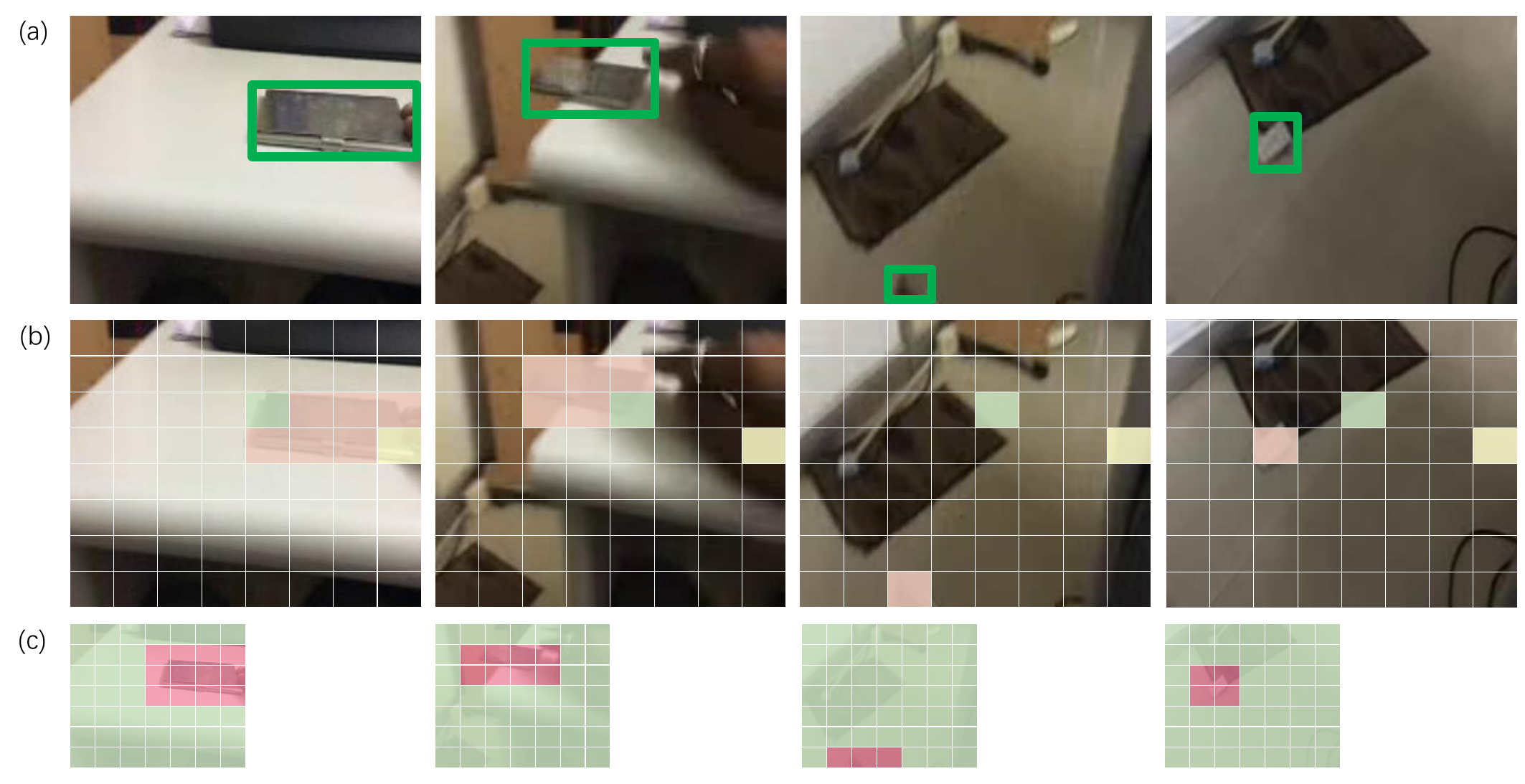}
  \caption{The differences between window-based attention and glance-like attention. \textbf{(a)} Frame sequences (\textit{pushing notebook so that it falls off the table}) from SSv2 \cite{goyal2017something} dataset. 
  \textbf{(b)} In window-based self-attention (e.g., Video-Swin \cite{liu2022video}), a square represents a \textbf{window} within which self-attention performs its calculations. When an object (marked with red) zooms, moves, and rotates across many different windows in frame sequences, the tokens in the yellow or green window of the first frame merely communicate with the tokens of the same object in subsequent frames at the later stage of the network.
  It is difficult to obtain an excellent receptive field to learn spatiotemporal information, particularly to model motion.  
  \textbf{(c)} After spatial downsampling, each square (which represents a \textbf{token}) can communicate freely with others, resulting in an attention mechanism that resembles a glance.
  }
  \label{fig:motivation}
      \vspace{-1em}
\end{figure*}

Some works focused on factorized \cite{bertasius2021space, arnab2021vivit} or window-based \cite{liu2022video, xiang2022spatiotemporal} self-attention to reduce the computational burden. Although these approaches partially resolved the computational cost problem of self-attention, they still sever the motion of regions of interest (ROI) in space and time, as illustrated in Figure~\ref{fig:motivation}. This limitation becomes more prominent, particularly when both the object and the camera are in motion. However, humans can quickly glance global temporal order of changes and track ROI in time and space. When examining object details, an additional gaze is needed to observe local spatial information such as shape, color, texture, etc. 


If we aim to make the model perform as efficiently as humans in video tasks, \textit{what should we consider for the temporal and spatial dimensions of videos?} We first acknowledge that sensory events are highly interdependent both in space and time \cite{attneave1954some}. Therefore, separating time and space \cite{neimark2021video, bertasius2021space, arnab2021vivit, liu2022video, xiang2022spatiotemporal} may impair the overall spatiotemporal features. However, does this mean that we should treat time and space equally, as did in previous methods \cite{carreira2017quo, qiu2017learning, bertasius2021space, neimark2021video, arnab2021vivit, liu2022video, li2022uniformer, li2021improved}?
In the human visual system, observation is sparse and attention is allocated over time through glance and gaze behavior \cite{large1999dynamics, deubel1996saccade}. This suggests that the human visual system places different emphasis on time and space at different time scales.

How can we apply this understanding to current action recognition settings and networks? In humans, after observing actions, memories are formed \cite{paller2002observing, stefan2005formation}, which make up visual perception. It is assumed that observation and memory occur across three time scales: long-term memory, clip-level view, and frame-level observation. This helps us understand the varying importance of time and space at different scales.

For \textbf{long-term-level memory} (referring to a single video containing multiple actions), people typically model the relationships between multiple actions, placing more emphasis on spatial information than temporal information, as shown in MeMViT \cite{wu2022memvit}. We believe this is because the interpretation or comprehension of individual actions has already been achieved \cite{paller2002observing}. For \textbf{clip-level view} (referring to a single video containing one action, as a part of a long-term-level memory, which is similar to most existing action recognition datasets), we should prioritize the temporal order as temporal information is crucial for agnostic actions and spatial information is more redundant with temporal information \cite{he2022masked, tong2022videomae}. Many human actions and interactions involve an ordered sequence of back-and-forth responses that cannot be modeled using a single or a few frames \cite{buch2022revisiting}. For \textbf{frame-level observation} (referring to a frame and its neighboring frames in a clip-level video), we should focus more on the local spatial information of ROI captured by a glance than on local temporal information, obtaining detailed information through gaze behavior.

The latter two levels should be taken into consideration. Humans store a bound trace of different action representations on time when forming memories \cite{paller2002observing}. However, current action recognition is based on clipped videos, where each video is within such temporal boundaries. In particular, during clip-level viewing, temporal information is crucial. Even if some actions that heavily rely on spatial information may appear to lack temporal information, this determination can only be made after viewing. Therefore, we reiterate that temporal information is invaluable when observing agnostic actions without prior knowledge. Furthermore, in this paper, the issue of the frame sampler \cite{korbar2019scsampler, wu2019adaframe, zhi2021mgsampler} is not discussed and instead only fixed stride \cite{carreira2017quo} or uniform \cite{goyal2017something} sampling is utilized, as it may disrupt the visual tempo and typically exists as a separate component without efficient communication with the backbone. 

\begin{figure*}
  \centering
  \includegraphics[width=\linewidth]{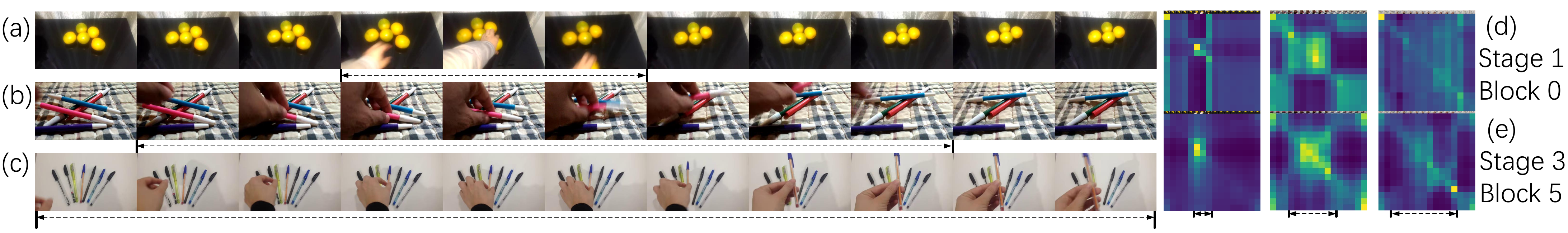}
    \vspace{-1em}
  \caption{By visualizing the similarity matrices and their corresponding three video sequences ((a) a fast tempo, (b) a slow tempo, (c) a slower tempo) from the same action category, we demonstrate that matrix $\boldsymbol{A}$ can effectively capture the tempo characteristics of the actions.}
  \label{fig:tempo}
      \vspace{-1em}
\end{figure*}

In summary, we propose a dual-path network called Overall Glance and Refined Gaze Transformer (shown in igure~\ref{fig:arc_over}), with Spatial-only Downsampling Attention (SoDA) in the Glance path to capture coarse-grained spatiotemporal information on clip level, and Masked Dynamic Convolution (MDConv) in the Gaze path to capture detailed local spatial or temporal information on frame level. We hope this can effectively leverage the advantages of the human visual system. 
In experiments, we find that a special balance between 2D and 3D convolution in Gaze path is required, rather than solely relying on 3D convolution as previous works \cite{carreira2017quo, qiu2017learning, tran2018closer, xie2018rethinking} did. We convert the token similarity matrix $\boldsymbol{A}$ in the Glance path into a frame similarity matrix, which can be seen as visual tempos \cite{feichtenhofer2019slowfast, yang2020temporal} as illustrated in Figure~\ref{fig:tempo}. This guides the balancing of 2D and 3D convolutions, allowing for adaptive handling of varying visual tempos through their combination. As shown in Table~\ref{tab:ab_frame}, imbalanced local temporal and spatial information leads to decreased performance. 
This work offers the following contributions: 

1) We rethink Transformer in action recognition and investigate the importance of time and space in video. Our proposed method incorporates the glance and gaze mechanisms, which mimic the human visual system. Although the glance and gaze mechanism has been applied in other fields \cite{yu2021glance, zhong2021glance, li2022glance}, to the best of our knowledge, there is no relevant research in action recognition, especially considering the difference between time and space in videos.

2) We propose Spatial-only Downsampling Attention (SoDA) to effectively process overall spatiotemporal information with low computational cost on clip level. This method allows for a similar function to the Glance mechanism, as the human visual system is sensitive to low-frequency content and motion. To the best of our knowledge, this is the first work to achieve competitive results using only coarse-grained overall self-attention in each network stage. Our method is not strictly a multi-scale approach, despite our consideration of multiple levels in time. 
Previous works have explored global or multi-scale attention, including \cite{li2021improved, wang2022deformable, li2022uniformer, yan2022multiview}, but they are either highly complex or rely on a fine-grained branch. In contrast, our approach achieves better performance than Video-Swin \cite{liu2022video} by $2.7\%$ on SSv2 using only coarse-grained overall self-attention, as demonstrated in Table~\ref{tab:ab_contribution}.

3) Compared to the state-of-the-art models, the proposed OG-ReG demonstrates remarkable performance on multiple benchmarks while also reducing computation costs. The model excels not only on spatial-heavy datasets (such as K400) but also shows substantial improvement on temporal-heavy datasets (such as SSv2).

\section{Related Work}\label{sec2}
\subsection{CNN-based Video Action Recognition} 
CNN-based video action recognition methods \cite{simonyan2014two, wang2016temporal, tran2015learning, carreira2017quo, qiu2017learning, tran2018closer, xie2018rethinking, feichtenhofer2020x3d} typically utilized two-stream networks, 3D convolution, or a combination thereof to extract appearance and motion information. As efficiency emerged as a significant concern, researchers began to investigate 2D convolution with temporal modeling \cite{lin2019tsm, feichtenhofer2019slowfast, kwon2020motionsqueeze}. Resently, 
MixRes~\cite{LU2024111686} combined hierarchical motion modeling and a Temporal Multiscale Motion Excitation (TMME) module to effectively capture short-term and long-term motion features.
MCNet~\cite{MUNSIF2024112480} integrated contextual visual and motion features through novel modules like the Temporal Optical Flow Learning Module (TOFLM) and Contextual Visual Features Learning Module (CVFLM) to achieve robust action recognition in dark environments.
GSLTA-CDFSAR~\cite{GUO2025113041} introduced a framework for cross-domain few-shot action recognition that aligns global sequences and local tuples using multiple-level distillation and domain-adaptive augmentation.
CANet~\cite{GAO2024111852} leveraged plug-and-play modules to effectively capture motion information across temporal, spatial, and channel dimensions in video-based action recognition. 

\subsection{Transformer-based Video Action Recognition} 
The exceptional performance of Transformer in image tasks has led to its introduction in video action recognition.
Several approaches, including VTN \cite{neimark2021video}, Timesformer \cite{bertasius2021space}, and ViViT \cite{arnab2021vivit} explored the factorization of the encoder, self-attention, and dot-product for videos. 
Video-Swin Transformer \cite{liu2022video} reduced computation by utilizing window-based attention. MViT \cite{fan2021multiscale, li2021improved} employed a pooling attention mechanism to reduce computation costs. Deformable Video Transformer \cite{wang2022deformable} leveraged motion cues to identify a sparse set of space-time locations to attend to for each query. Uniformer \cite{li2022uniformer} integrated local and global self-attention in a concise transformer format but considered one type of dependency, either global or local, in a single layer and sacrifices either global information during local modeling or vice versa \cite{si2022inception}.
TPS \cite{xiang2022spatiotemporal} had a similar computation and memory cost as 2D self-attention, achieved through patch shift operations. 
CLIP-MDMF~\cite{GUO2024112539} leveraged local and global temporal context extractors, probability prompt selectors, and multi-view mutual distillation to achieve superior accuracy and robustness by effectively fusing textual and visual modalities.
In addition, there are several studies focusing on co-learning and multimodal learning \cite{girdhar2022omnivore, ni2022expanding, piergiovanni2023rethinking, rasheed2023fine, xing2024emo, li2025deemo, xing2025emotionhallucer, wang2023internvid} that we have not compared with in this paper. The reason for this is that these methods employ a larger amount of training data and more robust training strategies.

\subsection{Reducing the Computation Cost of Self-Attention}
The computation cost of self-attention increases exponentially with the number of input tokens, making it intractable for visual tasks. To address the issue, several recent works focused on image tasks \cite{wang2021pyramid, wang2022pvt, zhang2021rest, zhang2022rest, fan2021multiscale, li2021improved} reduced the complexity of self-attention by combining it with key-value spatial reduction.
Other works \cite{liu2021swin, chu2021twins, huang2021shuffle} focused on window-based attention. However, the window-based methods may still limit the ability to capture long-range relations, especially for video inputs; the key-value spatial reduction methods may damage the spatial information.
MViTv2 \cite{li2021improved} not only has the aforementioned limitations, but also is insufficient for videos, especially for long-term action recognition and anticipation, and did not address the issue of self-attention's lack of high-frequency information.

\subsection{Combining Transformer and Convolution}
Many recent studies attempted to combine Transformer and convolution due to their complementary feature extraction properties. According to \cite{park2022vision, zhou2022understanding}, self-attention has weak inductive bias may impede training, and serves as a low-pass filter, with superior performance, generalization, and robustness compared to convolution. In contrast, convolution acts as a high-pass filter and remains essential. 
GG-Transformer \cite{yu2021glance} incorporated convolution in parallel with the value branch within self-attention. Uniformer \cite{li2022uniformer_image, li2022uniformer} used local attention in the early stages and global attention in the later stages. MixFormer \cite{chen2022mixformer} leveraged bi-directional interactions across branches to provide complementary information. Inception Transformer \cite{si2022inception} employed an Inception mixer to explicitly combine the benefits of self-attention, convolution, and max-pooling. Next-ViT \cite{li2022next} presented a cascaded network with self-attention and convolution, considering the latency and accuracy trade-off. However, most of the aforementioned methods have primarily focused on image tasks. To our knowledge, the integration of Transformer and convolution for video understanding has received relatively little attention, particularly in terms of considering the unique properties of video data. Uniformer \cite{li2022uniformer} used local attention to address ``local redundancy'', but had a vague definition of it, treated temporal and spatial information equally, and did not discuss the redundancy in time and space.

\subsection{Glance and Gaze}
Glance and Gaze \cite{deubel1996saccade} is a beneficial mechanism within human visual system \cite{chen2019drop, feichtenhofer2019slowfast, yu2021glance, min2022peripheral}. Individuals typically survey a scene quickly to identify and track ROI, and then examine the details of the ROI more closely. It extracts visual elementary features at different frequencies \cite{bullier2001integrated, bar2003cortical, kauffmann2014neural}, which echoes Transformer and convolution \cite{raghu2021vision, park2022vision}. Although glance and gaze have been studied in image recognition \cite{yu2021glance}, human-object interaction \cite{yu2021glance}, speech enhancement \cite{li2022glance}, and others, there is no relevant research that has considered it for action recognition.

\begin{figure*}
  \centering
  \includegraphics[width=0.95\textwidth]{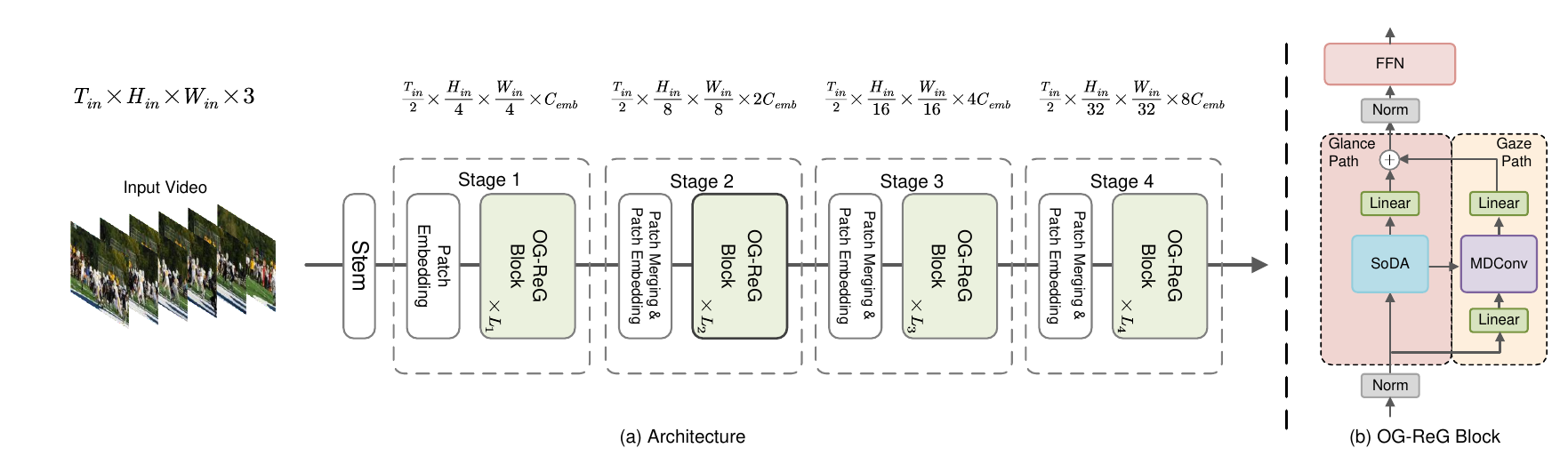}
  \vspace{-1em}
  \caption{The framework of OG-ReG Transformer. (a) An overview of the proposed Overall Glance and Refined Gaze (OG-ReG) Transformer.  (b)  The fundamental building block of OG-ReG, namely the OG-ReG block.}
  \label{fig:arc_over}
      \vspace{-1em}
\end{figure*}

\subsection{Spatiotemporal Redundancy in Videos} 
Spatiotemporal redundancy refers to the significant amount of redundant information between adjacent frames in both time and space in a video. Redundancy has been studied in various video tasks \cite{tang2007spatiotemporal, rav2006making, wei2009spatio, caballero2017real, li2018diversity, li2022uniformer}, but it is important to consider that redundancy differs in both time and space. Neglecting this may impede the design of efficient action recognition methods. On the one hand, temporal downsampling and image-centric baseline may encounter challenges in fine-grained action understanding and may be insufficient for inputs that require a deeper multi-frame understanding of event relationships or dynamics \cite{buch2022revisiting}. Therefore, spatiotemporal redundancy should not be defined only in the time domain. 
On the other hand, the ablation study in VideoMAE \cite{he2022masked, tong2022videomae} has shown that space-only/tube sampling with a higher masking ratio can work better than time-only/frame sampling with a lower ratio. This means that the spatial redundancy in videos is much greater than the temporal redundancy. The fact of the difference in redundancy between time and space in videos also supports our hypothesis.

\subsection{Visual Tempo} 
Visual tempo is used to characterize the dynamics and temporal scale of action \cite{feichtenhofer2019slowfast, yang2020temporal}. The complex temporal structure of action instances, particularly in terms of the various visual tempos, raises a challenge for action recognition. SlowFast \cite{feichtenhofer2019slowfast}, DTPN \cite{zhang2019dynamic}, and TPN \cite{yang2020temporal} focused on input-level frame pyramid that has level-wise frames sampled at different rates. Previous works only focused on the input visual tempo, without integrating with the human visual system or studying the balance of local temporal and spatial information.
\section{Methods}\label{sec3}
\subsection{OG-ReG Transformer}\label{subsec2}

Our aim is to design an effective attention mechanism that imitates human glance behavior to extract overall spatiotemporal information in videos, taking into account the varying importance and redundancy of temporal and spatial information.
Figure~\ref{fig:arc_over} illustrates the architecture of our Overall Glance and Refined Gaze Transformer. Additional details of model variants (OG-ReG- T, S, B) are provided in the Supplemental Materials.
We start with a video $ \mathcal{V} \in \mathbb{R}^{T_{in} \times H_{in} \times W_{in} \times 3}$, where $T_{in}$ represents the number of frames, and each frame has $H_{in} \times W_{in} \times 3$ pixels. Overlapped 3D patches as tokens are embedded by Conv Stem with total stride $2 \times 4 \times 4$. After patch embedding, we obtain input 3D tokens $Z_0 \in \mathbb{R}^{\frac{T_{in}}{2} \times \frac{H_{in}}{4} \times \frac{W_{in}}{4} \times{C_{emb}}}$.

A vanilla or hierarchical visual transformer consists of $L$ encoders with MSA, Layer-Norm (LN) and FFN. The transformer encoder could be represented as
\begin{equation}
\begin{aligned}
& \hat{\boldsymbol{Z}}_{l} = \operatorname{MSA} (\operatorname{LN} (\boldsymbol{Z}_{l-1} ) ) + \boldsymbol{Z}_{l-1}, \\
& \boldsymbol{Z}_{l}=\operatorname{FFN} (\operatorname{LN} (\hat{\boldsymbol{Z}}_{l} ) )+\hat{\boldsymbol{Z}}_{l},
\label{eq:Tra}
\end{aligned}
\end{equation}
\noindent where $\boldsymbol{Z}_{l-1}, \boldsymbol{Z}_{l} \in \mathbb{R}^{N \times d_m}$ denote the input and output features  the block $l$, respectively. To simplify the notation, the MSA operation is described in the setting of single-head attention and drop the layer superscripts of weights and neglect LN as follows:

\begin{equation}
\begin{aligned}
& \boldsymbol{Q}, \boldsymbol{K}, \boldsymbol{V} = \boldsymbol{Z}_{l-1} \boldsymbol{W}_{Q}, \boldsymbol{Z}_{l-1} \boldsymbol{W}_{K}, \boldsymbol{Z}_{l-1} \boldsymbol{W}_{V}, \\
& \hat{\boldsymbol{Z}}_{l} = \operatorname{SoftMax} (\boldsymbol{Q} \boldsymbol{K}^{T} / \sqrt{d} ) \boldsymbol{V},
\label{eq:MSA}
\end{aligned}
\end{equation}

\noindent where $\boldsymbol{W}_{Q}$, $\boldsymbol{W}_{K}$, $\boldsymbol{W}_{V}$ are the weights of linear projection, and $\boldsymbol{Q}, \boldsymbol{K}, \boldsymbol{V}$ represent the $query$, $key$ and $value$. Our network architecture fundamentally adheres to the design principles of a hierarchical vision transformer. The essential distinction lies in our utilization of SoDA in conjunction with MDConv to emulate the mechanisms of human visual perception.

\subsection{Overall Glance Path} 

\begin{figure*}
  \centering
  \includegraphics[width=0.9\linewidth]{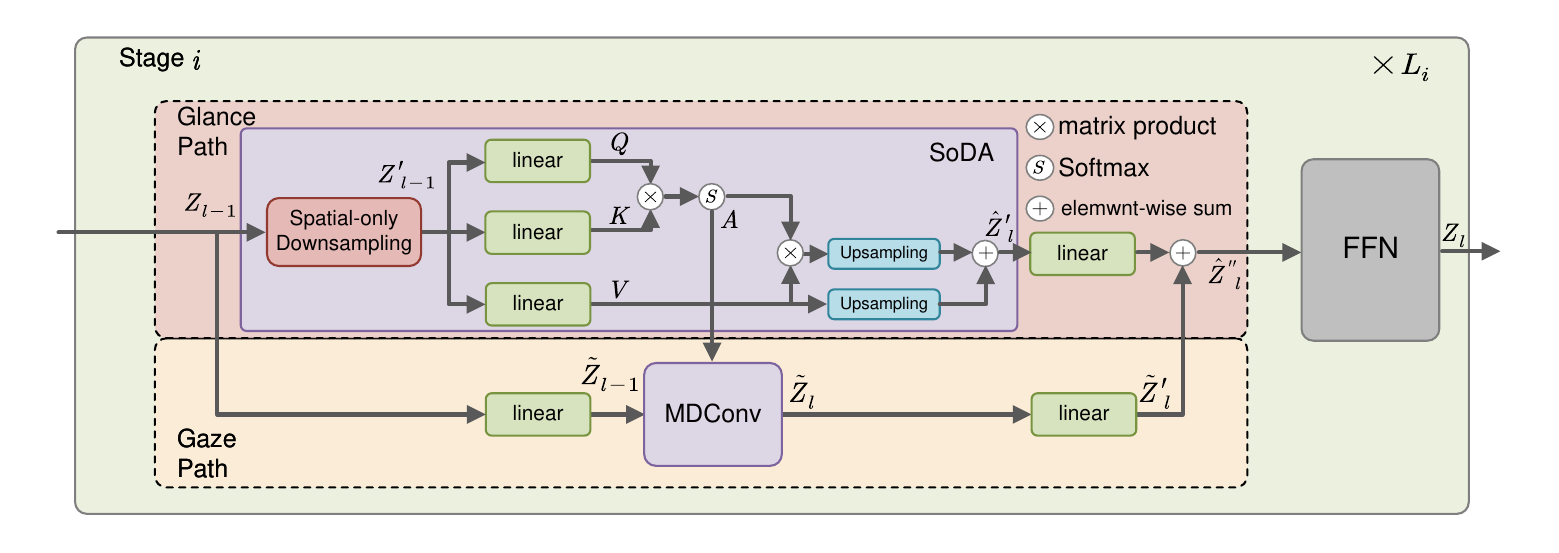}
  \vspace{-1em}
  \caption{Details of the OG-ReG block (neglecting Layer-Norm).}
  \label{fig:soda}
      \vspace{-1em}
\end{figure*}

To effectively extract overall spatiotemporal features, we propose the Spatial-only Downsampling Attention (SoDA), which is based on the principle of reducing redundant spatiotemporal information as discussed in Section~\ref{sec2}. This allows for a more coarse-grained self-attention to be implemented for video tasks. In layer $l$, the input tokens $\boldsymbol{Z}_{l-1} \in \mathbb{R}^{N \times C}$, where $N = T_{l-1} \times H_{l-1} \times W_{l-1}$. Considering the redundancy of spatial information in videos, we  perform a downsampling operation exclusively on the spatial dimensions of the input tokens (as illustrated in Figure~\ref{fig:soda}). This approach not only significantly reduces computational costs but also achieves commendable results for action recognition, as demonstrated in Table~\ref{tab:ab_contribution}. The formulation of SoDA (a single head) in layer $l$ can be expressed as
\begin{equation}
\begin{aligned}
& \boldsymbol{Z}_{l-1}^{\prime} =  \operatorname{DR} ( \boldsymbol{Z}_{l-1} ),  \\
& \boldsymbol{Q}, \boldsymbol{K}, \boldsymbol{V} = \boldsymbol{Z}_{l-1}^{\prime}\boldsymbol{W}_{Q}, \boldsymbol{Z}_{l-1}^{\prime}\boldsymbol{W}_{K}, \boldsymbol{Z}_{l-1}^{\prime}\boldsymbol{W}_{V}, \\
& \hat{\boldsymbol{Z}_{l}} = \boldsymbol{A} \boldsymbol{V}, \boldsymbol{A} = \operatorname{SoftMax} (\boldsymbol{Q} \boldsymbol{K}^{T} / \sqrt{d} ), \\
& \hat{\boldsymbol{Z}_{l}^{\prime}} = \operatorname{UR} ( \hat{\boldsymbol{Z}}_{l} ) + \operatorname{UR} ( \boldsymbol{V} ), 
\label{eq:SDSA}
\end{aligned}
\end{equation}

\noindent Where $\boldsymbol{W}_{Q}$, $\boldsymbol{W}_{K}$, $\boldsymbol{W}_{V}$ are linear projections and $d$ is the hidden dimension in self-attention. The operations for reducing and restoring the spatial resolution of the input are denoted as $\operatorname{DR}(\cdot)$ and $\operatorname{UR}(\cdot)$, respectively, and can be defined as follows:
\begin{equation}
\begin{aligned}
& \operatorname{DR}(\boldsymbol{X}) = \operatorname{ReS} ( \operatorname{Down} ( \operatorname{ReS} (\boldsymbol{X}),  s)), \\
& \operatorname{UR}(\boldsymbol{X}) = \operatorname{ReS} ( \operatorname{Up} ( \operatorname{ReS} (\boldsymbol{X}),  s)). \\
\label{eq:SRUR}
\end{aligned}
\end{equation}
\noindent Here, $\boldsymbol{X} \in \mathbb{R}^{N \times{C}}$ represents an input, where $s$ represents the downsample/upsample ratio, typically is set to $[8, 4, 2, 1]$ at different stages. The operations of reducing/restoring the spatial resolution of the input only. $\operatorname{ReS} (\boldsymbol{X})$ reshapes the input sequence of $\boldsymbol{X} \in \mathbb{R}^{N \times{C}}$ to $\mathbb{R}^{T \times H \times W \times {C}}$ or back. $\operatorname{Down}(\boldsymbol{x}, s)$ and $\operatorname{Up}(\boldsymbol{X}, s)$ downsample the feature $\boldsymbol{X} \in \mathbb{R}^{T \times H \times W \times {C}}$ to $ \mathbb{R}^{T \times \frac{H}{s} \times \frac{W}{s} \times {C}}$, or upsample back. 

\begin{figure}
  \centering
  \includegraphics[width=\linewidth]{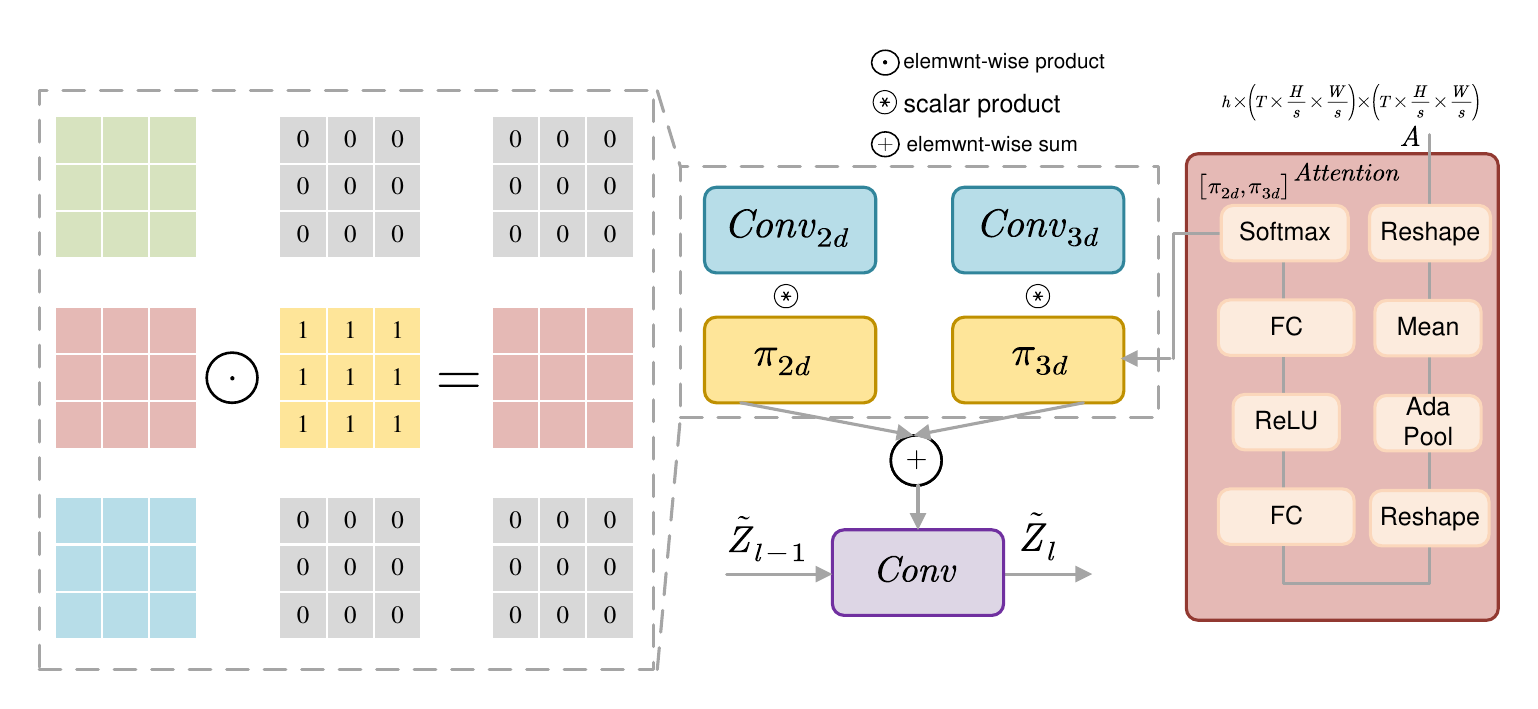}
    \vspace{-2em}
  \caption{Details of the MDConv.}
  \label{fig:mdconv}
      \vspace{-1em}
\end{figure}

\subsection{Refined Gaze Path} 
To extract local spatiotemporal information, we discovered that relying solely on 3D convolutions, as in previous works \cite{carreira2017quo, qiu2017learning, tran2018closer, xie2018rethinking} did, is inadequate. Instead, it is necessary to modulate 2D and 3D convolutions differently for time and space.
To achieve this, we convert token similarity matrix $ \boldsymbol{A}$ from the Glance path to the frame similarity matrix to guide the balance of local temporal and spatial information. We introduce the MDConv, which integrates 2D and 3D convolution (refer to the bottom of Figure~\ref{fig:arc_over} for details). Specifically, given the input after linear projection ${\boldsymbol{\tilde{Z}}_{l-1}} \in \mathbb{R}^{T_{l-1} \times H_{l-1} \times W_{l-1} \times {C_{l-1}}}$, the MDConv is formulated as follows:
\begin{equation}
\begin{aligned}
& \pi_{2d}, \pi_{3d} = \operatorname{Attention} (\boldsymbol{A}), \\
& \operatorname{ s.t. }~ 0 \le \pi_{2d}, \pi_{3d} \le 1, \pi_{2d} + \pi_{3d} = 1, \\
& \boldsymbol{{W}}_{2d+3d} = \pi_{2d} \boldsymbol{{W}}_{2d} + \pi_{3d} \boldsymbol{{W}}_{3d}, \\
& \boldsymbol{\tilde{Z}}_l = \operatorname{MDConv}(\boldsymbol{\tilde{Z}}_{l-1}, \boldsymbol{A}) = \boldsymbol{\tilde{Z}}_{l-1} \boldsymbol{{W}}_{2d+3d} +\boldsymbol{b} , \\
\label{eq:MDC}
\end{aligned}
\end{equation}
\noindent where $\operatorname{Attention}(\cdot)$ converts the token similarity matrix $ \boldsymbol{A}$ to the modulation factors $[\pi_{2d}, \pi_{3d}]$ (shown in Figure~\ref{fig:mdconv}). $\boldsymbol{W}_{2d}$, $\boldsymbol{W}_{3d}$ denote the weights, and $\boldsymbol{b}$ represents the biases. By reshaping and averaging the matrix we obtain $\boldsymbol{A}^{\prime} \in \mathbb{R}^{T \times T }$, which represents the frame similarity (see Figure~\ref{fig:tempo}). To handle different frame sampling rates during inference ($T^{\prime}$) from training ($T$, typically $T = T'$), as the shape of the FC parameters (related to $T$) is fixed, we add an adaptive pooling layer to adjust the matrix to match the training shape. Finally, we obtain the modulation factors $[\pi_{2d}, \pi_{3d}]$ for the 2D and 3D kernels.

To add $\boldsymbol{W}_{2d}$ with $\boldsymbol{W}_{3d}$ in PyTorch, we mask the temporal dimension of 3D kernel to obtain a 2D Conv. Let $\boldsymbol{\tilde{W}}_{3d} \in \mathbb{R}^{kd \times kh \times kw}$ be a 3D convolution kernel, where $kd$, $kh$ and $kw$ are the dimension of time, height and weight. We perform element-wise production between the kernel and the fixed mask $\boldsymbol{M} \in \mathbb{R}^{kd \times kh \times kw}$ to obtain the 2D convolution kernel with a 3D shape: $\boldsymbol{\tilde{W}}_{2d} = \boldsymbol{M} \odot \boldsymbol{\tilde{W}}_{3d}$. To extract information of the central frame, we set $\boldsymbol{M}[\frac{kd-1}{2}, :, :] = 1$ for central time elements and 0 for the others. This process functionally transforms a 3D kernel into a 2D one. Through MDConv, the Gaze path captures local high-frequency information and complements the features obtained by the Glance path. 
Although the method of modulating convolution kernels has also been used in Dynamic Convolution \cite{chen2020dynamic}, it only focused on image tasks, while we consider the characteristics of videos when designing mask and attention mechanisms.
The two paths and their interactions mimic that humans quickly glance at the global temporal order of changes and track ROI in time and space, and when examining object details, an additional gaze is needed to observe local spatial and local temporal information.

\begin{table*}
  \caption{Comparison with state-of-the-art on K400.}
  \label{tab:k400cmp}
  \begin{tabular*}{\textwidth}{l@{\extracolsep{\fill}}c@{\extracolsep{\fill}}cc@{\extracolsep{\fill}}c@{\extracolsep{\fill}}cc}
    \toprule
    Method & Pretrain & Top-1 & Top-5 & Views & FLOPs & Param \\
    \midrule
    R(2+1)D \cite{tran2018closer} & - & 72.0 & 90.0 & $ 10 \times 1 $ & 75  & 61.8 \\
    I3D \cite{carreira2017quo} & ImageNet-1K & 72.1 & 90.3 & $ 1 \times 1 $ & 108  & 28.0 \\
    NL-I3D \cite{wang2018non}& ImageNet-1K & 77.7 & 93.3 & $ 6 \times 10 $ & 32  & 35.3 \\
    CoST \cite{li2019collaborative} & ImageNet-1K & 77.5 & 93.2 & $ 3 \times 10 $ & 33  & 35.3 \\
    SlowFast-R50 \cite{feichtenhofer2019slowfast} & ImageNet-1K & 75.6 & 92.1 & $ 3 \times 10 $ & 36  & 32.4 \\
    X3D-XL \cite{feichtenhofer2020x3d} & - & 79.1 & 93.9 & $ 3 \times 10 $ & 48  & 11.0 \\
    TSM \cite{lin2019tsm} & ImageNet-1K & 74.7 & 91.4 & $ 3 \times 10 $ & 65  & 24.3 \\
    TEINet \cite{liu2020teinet} & ImageNet-1K & 76.2 & 92.5 & $ 3 \times 10 $ & 66  & 30.4 \\
    TEA \cite{li2020tea} & ImageNet-1K & 76.1 & 92.5 & $ 3 \times 10 $ & 70  & 24.3 \\
    TDN \cite{wang2021tdn} & ImageNet-1K & 77.5 & 93.2 & $ 3 \times 10 $ & 72  & 24.8 \\
    \midrule
    Timesformer-L \cite{bertasius2021space} & ImageNet-21K & 80.7 & 94.7 & $ 1 \times 3 $ & 2380 & 121.4 \\
    
    ViT-B-VTN \cite{neimark2021video} & ImageNet-21K & 78.6 & 93.7 & $ 1 \times 1 $ & 4218 & 114.0 \\
    
    ViViT-L/16x2 320 \cite{arnab2021vivit} & ImageNet-21K & 81.3 & 94.7 & $ 4 \times 3 $ & 3992 & 310.8 \\
    
    DVT-B \cite{wang2022deformable} & ImageNet-21K & 81.5 & 95.2 & $ 1 \times 5 $ & 128 & - \\
    
    UniFormer-B, 32 \cite{li2022uniformer} & ImageNet-1K & 82.9 & 95.4 & $ 4 \times 1 $ & 256 & - \\

    MViTv2-B, 32 \cite{li2021improved} &   & 82.9 & 95.7 & $ 1 \times 5 $ & 225 & 51.2 \\


    Video-Swin-T \cite{liu2022video} & ImageNet-1K & 78.8 & 93.6 & $ 4 \times 3 $ & 88 & 28.2 \\
    Video-Swin-S \cite{liu2022video} & ImageNet-1K & 80.6 & 94.5 & $ 4 \times 3 $ & 166 & 49.8 \\
    Video-Swin-B \cite{liu2022video} & ImageNet-1K & 80.6 & 94.6 & $ 4 \times 3 $ & 282 & 88.1 \\
    Video-Swin-B \cite{liu2022video} & ImageNet-21K & 82.7 & 95.5 & $ 4 \times 3 $ & 282 & 88.1 \\
    
    PST-T \cite{xiang2022spatiotemporal} & ImageNet-1K & 78.2 & 92.2 & $ 4 \times 3 $ & 72 & 28.5 \\
    PST-B \cite{xiang2022spatiotemporal} & ImageNet-21K & 81.8 & 95.4 & $ 4 \times 3 $ & 247 & 88.8 \\
    
    \midrule
    OG-ReG-T & ImageNet-1K & 79.5 & 94.2 & $ 4 \times 3 $ & 73 & 32.4 \\
    OG-ReG-S & ImageNet-1K & 80.8 & 94.7 & $ 4 \times 3 $ & 139 & 61.9 \\
    OG-ReG-B & ImageNet-1K & 81.0 & 94.9 & $ 4 \times 3 $ & 245 & 96.6 \\
    OG-ReG-B & ImageNet-21K & {\bf 83.0} & 95.7 & $ 4 \times 3 $ & 245 & 96.6 \\
    \bottomrule
  \end{tabular*}
      \vspace{-1em}
\end{table*}

\section{Experiments}\label{sec4}

\subsection{Datasets}\label{subsec3}

The Kinetics-400 (K400) \cite{carreira2017quo} is a widely used, large-scale, spatial-heavy action recognition dataset comprised of approximately 240k training video clips and 20k validation video clips in 400 categories of human actions. The video clips are collected from YouTube and trimmed to about 10 seconds. The Something-Something v2 (SSv2) \cite{goyal2017something} is a large-scale, temporal-heavy dataset of daily actions that focuses on object motion, without differentiating manipulated objects. It contains approximately 221k video clips and 174 classes. The Diving-48 v2 dataset \cite{li2018resound} is a fine-grained video dataset of competitive diving and includes 18k trimmed video clips of 48 specific dive actions. 

\subsection{Experiment setup}\label{subsec4}
\textbf{Training} We follow the data processing approach and training strategies of Video-Swin \cite{liu2022video} and PST \cite{xiang2022spatiotemporal} for all datasets. During training, we resize the short side of raw images to 256, then apply random resized crop, random flip, and AutoAugment for augmentation and employ an AdamW optimizer for 30 epochs using a cosine decay learning rate scheduler and 2.5 epochs of linear warmup. The backbone is initialized from the model pretrained on ImageNet-1K for K400 and Diving-48. For SSv2 and Diving-48, we use the K400 to accomplish pretraining. Unless specified otherwise, for all model variants, 32 frames are sampled from each full-length video by using a temporal stride of 2 for K400 and adopting a uniform random sampling strategy for SSv2 and Diving-48. For OG-ReG-T, the base learning rate, stochastic depth rate, weight decay, and batch size are set to $10^{-3}$, 0.1, 0.02, and 64, respectively. For the larger model OG-ReG-B, learning rate, drop path rate, and weight decay is set to $3 \times 10^{-4}$, 0.3, 0.05, respectively. For detailed training and pretraining settings, please refer to the supplementary material.

\begin{table*}
  \caption{Comparison with state-of-the-art on SSv2.}
  \label{tab:ssv2cmp}
  \begin{tabular*}{\textwidth}{l@{\extracolsep{\fill}}c@{\extracolsep{\fill}}cc@{\extracolsep{\fill}}c@{\extracolsep{\fill}}cc}
    \toprule
    Method & Pretrain & Top-1 & Top-5 & Views & FLOPs & Param \\
    \midrule
    TSM \cite{lin2019tsm} & K400 & 63.4 & 88.5 & $ 3 \times 2 $ & 65 & 24.3 \\
    TEINet \cite{liu2020teinet} & ImageNet-1K & 62.1 & - & $ 1 \times 1 $ & 66 & 30.4 \\
    TDN \cite{wang2021tdn} & ImageNet-1K & 65.3 & 89.5 & $ 1 \times 1 $ & 72 & 24.8 \\
    ACTION-Net \cite{wang2021action} & ImageNet-1K & 64.0 & 89.3 & $ 1 \times 1 $ & 70 & 28.1 \\
    SlowFast-R101, 8x8 \cite{feichtenhofer2019slowfast} & K400 & 63.1 & 87.6 & $ 3 \times 1 $ & 106 & 53.3 \\
    MSNet \cite{kwon2020motionsqueeze} & ImageNet-1K & 64.7 & 89.4 & $ 1 \times 1 $ & 101 & 24.6 \\
    blVNet \cite{fan2019more} & ImageNet-1K & 65.2 & 90.3 & $ 1 \times 1 $ & 129 & 40.2 \\
    \midrule
    Timesformer-HR \cite{bertasius2021space} & ImageNet-21K & 62.5 & - & $ 3 \times 1 $ & 1703 & 121.4 \\
    ViViT-L/16x2 \cite{arnab2021vivit} & ImageNet-21K & 65.9 & 89.9 & $ 3 \times 1 $ & 903 & 352.1 \\
    Mformer-L \cite{patrick2021keeping} & ImageNet-21K+K400 & 68.1 & 91.2 & $ 3 \times 1 $ & 1185 & 86 \\
    X-ViT \cite{bulat2021space} & ImageNet-21K+K400 & 66.2 & 90.6 & $ 3 \times 1 $ & 283 & 92 \\
    SIFAR-L \cite{fan2021image} & ImageNet-21K+K400 & 64.2 & 88.4 & $ 3 \times 1 $ & 576 & 196 \\
    
    DVT-B \cite{wang2022deformable} & ImageNet-21K+K400 & 68.0 & 91.0 & $ 3 \times 1 $ & 128 & - \\

    UniFormer-B, 32 \cite{li2022uniformer} & ImageNet-1K + K400 & 71.2 & 92.8 & $ 3 \times 1 $ & 259 & - \\

    MViTv2-B, 32 \cite{li2021improved} & K400 & 70.5 & 92.7 & $ 3 \times 1 $ & 225 & 51.1 \\


    Video-Swin-T \cite{liu2022video} & - & 66.2 & 90.8 & $ 3 \times 1 $ & - & - \\
    Video-Swin-B \cite{liu2022video} & ImageNet-21K+K400 & 69.6 & 92.7 & $ 3 \times 1 $ & 321 & 88.1 \\
    
    PST-T \cite{xiang2022spatiotemporal} & ImageNet-1K+K400 & 67.3 & 90.5 & $ 3 \times 1 $ & 72 & 28.5 \\
    PST-B \cite{xiang2022spatiotemporal} & ImageNet-21K+K400 & 69.2 & 91.9 & $ 3 \times 1 $ & 247 & 88.8 \\
    
    \midrule
    OG-ReG-T & ImageNet-1K+K400 & 68.9 & 91.2 & $ 3 \times 1 $ & 73 & 32.4 \\
    OG-ReG-S & ImageNet-1K+K400 & 70.4 & 92.3 & $ 3 \times 1 $ & 139 & 61.9 \\
    OG-ReG-B & ImageNet-21K+K400 & \textbf{71.7} & 93.1 & $ 3 \times 1 $ & 245 & 96.6 \\
    \bottomrule
  \end{tabular*}
      \vspace{-1em}
\end{table*}

\begin{table*}
  \caption{Comparison with state-of-the-art on Diving-48 v2.}
  \label{tab:divv2cmp}
  \begin{tabular*}{\textwidth}{l@{\extracolsep{\fill}}c@{\extracolsep{\fill}}cc@{\extracolsep{\fill}}c@{\extracolsep{\fill}}cc}
    \toprule
    Method & Pretrain & Top-1 & Top-5 & Views & FLOPs & Param \\
    \midrule
    SlowFast-R101, 8x8 \cite{feichtenhofer2019slowfast} & K400 & 77.6 & - & $ 3 \times 1 $ & 106 & 53.3 \\
    \midrule
    Timesformer \cite{bertasius2021space} & ImageNet-21K & 74.9 & - & $ 3 \times 1 $ & 196 & 121.4 \\
    Timesformer-HR \cite{bertasius2021space} & ImageNet-21K & 78.0 & - & $ 3 \times 1 $ & 1703 & 121.4 \\
    Timesformer-L \cite{bertasius2021space} & ImageNet-21K & 81.0 & - & $ 3 \times 1 $ & 2380 & 121.4 \\
    
    DVT-B \cite{wang2022deformable} & ImageNet-1K & 86.0 & - & - & 128 & - \\
    
    PST-T \cite{xiang2022spatiotemporal} & ImageNet-1K & 79.2 & 98.2 & $ 3 \times 1 $ & 72 & 28.5 \\
    PST-T \cite{xiang2022spatiotemporal} & ImageNet-1K+K400 & 81.2 & 98.7 & $ 3 \times 1 $ & 72 & 28.5 \\
    PST-B \cite{xiang2022spatiotemporal} & ImageNet-21K & 83.6 & 98.5 & $ 3 \times 1 $ & 247 & 88.1 \\
    PST-B \cite{xiang2022spatiotemporal} & ImageNet-21K+K400 & 85.0 & 98.6 & $ 3 \times 1 $ & 247 & 88.1 \\
    
    \midrule
    OG-ReG-T & ImageNet-1K & 83.9 & 98.1 & $ 3 \times 1 $ & 73 & 32.4 \\
    OG-ReG-T & ImageNet-1K+K400 & 84.9 & 98.7 & $ 3 \times 1 $ & 139 & 61.9 \\
    OG-ReG-B & ImageNet-21K & 87.0 & 98.5 & $ 3 \times 1 $ & 245 & 96.6 \\
    OG-ReG-B & ImageNet-21K+K400 & 88.1 & 98.9 & $ 3 \times 1 $ & 245 & 96.6 \\
    \bottomrule
  \end{tabular*}
      \vspace{-1em}
\end{table*}

\textbf{Testing} 
We adopt the testing strategy in SOTA methods \cite{arnab2021vivit, liu2022video, xiang2022spatiotemporal} for a fair comparison. In addition, we use the dense sampling strategy \cite{arnab2021vivit} with 4 views and three-crop during inference on K400. For SSv2 and Diving-48, we adopt uniform sampling and three-crop testing.

\subsection{Comparison with SOTA}
\textbf{K400.} Table~\ref{tab:k400cmp} shows comparisons with the SOTA in terms of the pretrained dataset, classification results, inference protocol, FLOPs, and parameter numbers, including both convolution-based and Transformer-based on K400. OG-ReG-T achieves $ 79.5\% $ top-1 accuracy and outperforms the majority of CNN-based methods with fewer total FLOPs. Compared to Transformer-based methods, our OG-ReG-B achieves $ 73.0\% $ with less computation cost. With a fair comparison to Video-Swin \cite{liu2022video} and PST \cite{xiang2022spatiotemporal}, our OG-ReG outperforms them on different sizes and pretrained datasets. Our OG-ReG-T outperforms Video-Swin-T \cite{liu2022video} by $0.7\%$ and PST-T \cite{xiang2022spatiotemporal} by $ 1.3\% $.

\textbf{SSv2.} The comparison of SOTA with our approach on SSv2 is shown in Table~\ref{tab:ssv2cmp}. Our OG-ReG-T obtains $68.9\%$ top-1 accuracy, which outperforms all the CNN-based methods with similar FLOPs and outperforms most Transformer-based methods with larger FLOPs and pretrained datasets. Specifically, OG-ReG-T outperforms Video-Swin-T by $2.7\%$ and PST-T \cite{xiang2022spatiotemporal} by $1.6\%$. OG-ReG-S pretrained on ImageNet-1K+K400 outperforms Video-Swin-B \cite{liu2022video} and PST-B \cite{xiang2022spatiotemporal} pretrained on ImageNet-21K+K400 by $ 0.6\% $ and $ 1.0\% $. OG-ReG-B pretrained on ImageNet-21K+K400 outperforms Video-Swin-B \cite{liu2022video} and PST-B \cite{xiang2022spatiotemporal} by $ 2.1\% $ and $ 2.5\% $. The performance of the OG-ReG family on SSv2 demonstrates its remarkable temporal modeling ability.


\textbf{Diving-48 v2.} In Tables~\ref{tab:divv2cmp}, we further report the performances of OG-ReG on Diving-48 v2. The experimental results confirm that our method has excellent capability on fine-grained action dataset.

\subsection{Ablation study}
To verify our hypothesis, a series of experiments are conducted in this subsection, using OG-ReG-T (pretrained on ImageNet-1K) on K400 and SSv2 datasets.

\begin{table*}
  \caption{Ablation study on Clip Level on K400 and SSv2.}
  \label{tab:ab_clip}
  \begin{tabular}{ccccccc}
    \toprule
    \multirow{2}{*}[-0.3em]{Spatial} & \multirow{2}{*}[-0.3em]{Temporal} & \multirow{2}{*}[-0.3em]{Pixel Shuffle} & \multicolumn{2}{c}{ K400 } & \multicolumn{2}{c}{ SSv2 } \\ \cmidrule(lr){4-5} \cmidrule(lr){6-7}%
     & & & Top-1 & Top-5 & Top-1 & Top-5 \\
    \midrule
    2D & w/o & 2D & {\bf 79.5} & {\bf 94.2} & {\bf 68.9} & {\bf 91.2} \\
    \multicolumn{2}{c}{ 3D } & 3D & 78.6 & 93.5 & 68.2 & 90.9 \\
    2D & 1D & R(2+1)D & 76.2 & 92.5 & 66.3 & 90.0 \\
    \bottomrule
  \end{tabular}
\end{table*}

\textbf{Clip Level.} 
We begin with a straightforward experiment to verify our hypothesis stated in the Introduction: temporal information is crucial for action recognition on clip level. As shown in Table~\ref{tab:ab_clip}, downsampling the temporal dimension with 3D convolution leads to a $0.9\%$ top-1 accuracy decrease on K400 and a $0.7\%$ decrease on SSv2. This could be attributed to two reasons: first, the temporal dimension is crucial to self-attention and should not be downsampled; second, downsampling and pixel shuffle methods are inappropriate. To further investigate the latter issue, we also attempted to use R(2+1)D convolution to replace the 3D operation. As seen in Table~\ref{tab:ab_clip}, both in 3D and R(2+1)D settings, there is a significant drop, indicating that temporal downsampling is not appropriate for self-attention on clip level for video action recognition.

\begin{table}
  \centering
  \caption{Ablation study on Frame Level pattern on K400 and SSv2.}
  \label{tab:ab_frame}
  \begin{tabular}{lcccc}
    \toprule
    \multirow{2}{*}[-0.3em]{ Pattern } & \multicolumn{2}{c}{ K400 } & \multicolumn{2}{c}{ SSv2 } \\ \cmidrule(lr){2-3} \cmidrule(lr){4-5}%
     & Top-1 & Top-5 & Top-1 & Top-5 \\ 
    \midrule
    2D & 79.2 & 94.0 & 68.5 & 91.3 \\
    3D & 79.1 & 93.8 & 68.6 & 91.2 \\
    fixed 2D + 3D & 79.2 & 94.0 & 68.5 & 91.2 \\
    factorized 3D & 79.3 & 93.7 & 68.6 & 91.2 \\
    2D+3D & {\bf79.5} & {\bf94.2} & \bf{68.9} & 91.2 \\
    \bottomrule
  \end{tabular}
\end{table}

\textbf{Frame Level.} 
We further studied the importance of local temporal and spatial information, based on the effective modeling of overall spatiotemporal information.
We experimented with different MDConv patterns on various datasets, including 2D, 3D, fixed 2D+3D ($\pi_{2d} = \pi_{3d} = 0.5$), factorized 3D, and 2D+3D. The results are presented in Table~\ref{tab:ab_frame}. 
Comparing the 2D+3D and variants revealed that balancing local temporal and spatial information can achieve the best performance. This may be because the overall spatiotemporal attention can more effectively extract temporal information, while coordinating the extraction of local spatiotemporal information requires the modulation of 2D and 3D information. 

\begin{table}
\centering
  \caption{Ablation study on Contributions of the Two Paths on K400 and SSv2.}
  \label{tab:ab_contribution}
  \begin{tabular}{cccccc}
    \toprule
    \multirow{2}{*}[-0.3em]{SoDA} & \multirow{2}{*}[-0.3em]{MDConv} & \multicolumn{2}{c}{ K400 } & \multicolumn{2}{c}{ SSv2 } \\ \cmidrule(lr){3-4} \cmidrule(lr){5-6}%
     & & Top-1 & Top-5 & Top-1 & Top-5 \\
    \midrule
    \checkmark & & 78.5 & 93.6 & 68.2 & 90.9 \\
     & \checkmark & 77.0 & 92.7 & 66.0 & 90.0 \\
    \checkmark & \checkmark & {\bf 79.5} & {\bf 94.2} & {\bf 68.9} & {\bf 91.2} \\
    \midrule
    \multicolumn{2}{c}{Video-Swin-T \cite{liu2022video}} & 78.8 & 93.6 & 66.2 & 90.8 \\
    \multicolumn{2}{c}{PST-T \cite{xiang2022spatiotemporal}} & 78.2 & 92.2 & 67.3 & 90.5 \\
    \bottomrule
  \end{tabular}
\end{table}

\begin{figure}
  \centering
  \includegraphics[width=\linewidth]{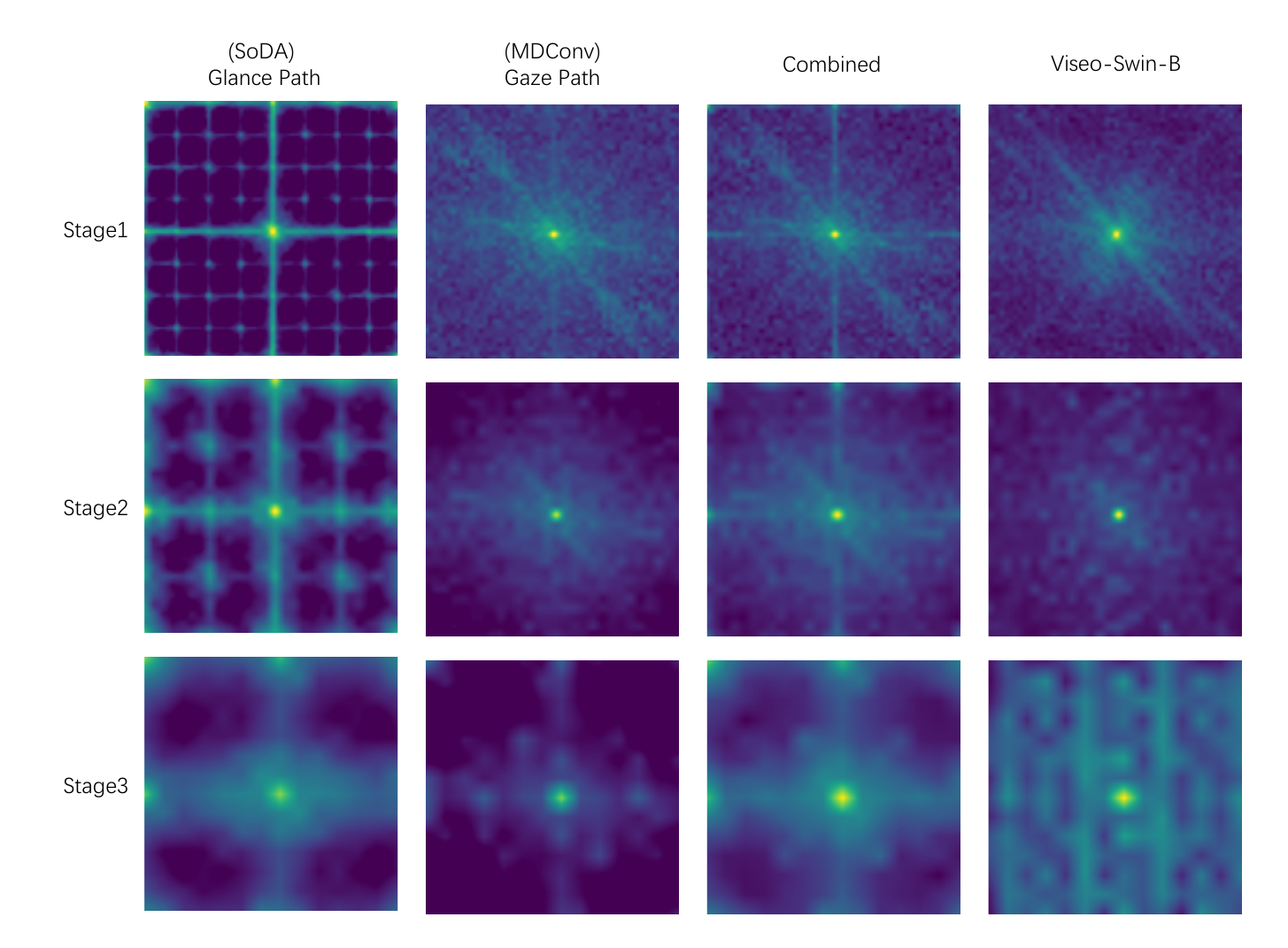}
    \vspace{-1em}
  \caption{Fourier spectrum of features in OG-ReG-T and Video-Swin-B.}
  \label{fig:fourier}
      \vspace{-1em}
\end{figure}

\textbf{Contributions of the Two Paths.} 
We further investigated the specific contributions of the two paths. As shown in Table~\ref{tab:ab_contribution}, combining the two paths improved the performance. Our SoDA achieved $0.3\%$ lower accuracy on K400 compared to Video-Swin-T, but $2\%$ higher on SSv2, demonstrating that SoDA pays more attention to temporal information. MDConv has a competitive performance with Video-Swin on SSv2. These results indicate that window-based attention disrupts the overall spatiotemporal information, especially temporal information.
The experiment also confirms that SoDA tends to the low-frequency characteristic and MDConv tends to the high-frequency characteristic, as shown by the Fourier spectrum of OG-ReG-T in Figure~\ref{fig:fourier}. Despite downsampling the input, SoDA can still learn sufficient low-frequency information. Furthermore, as the network stage gets deeper, SoDA plays an increasingly important role, occupying more components. In addition, it can be observed that in the early stage of the network, there are sufficient spectral components and no significant differences between Video-Swin-B and ours can be seen. However, in the later stage of the network, our method exhibits clearer spectral features, while Video-Swin-B appears to have some noise. The above ablation experiments verified our hypothesis in the Introduction. 


\begin{table}
\centering
  \caption{Ablation study on Guide Manner on K400 and SSv2.}
  \label{tab:ab_guide}
  \begin{tabular}{lcccc}
    \toprule
    \multirow{2}{*}[-0.3em]{Manner} & \multicolumn{2}{c}{ K400 } & \multicolumn{2}{c}{ SSv2 } \\ \cmidrule(lr){2-3} \cmidrule(lr){4-5}%
     & Top-1 & Top-5 & Top-1 & Top-5 \\
    \midrule
    Dynamic Conv \cite{chen2020dynamic} & 79.3 & 94.1 & 68.5 & 91.3 \\
    Ours & \textbf{79.5} & \textbf{94.2} & \textbf{68.9} & 91.2 \\
    \bottomrule
  \end{tabular}
\end{table}

\textbf{Guiding Manner.} 
As shown in Table~\ref{tab:ab_guide}, we investigated the difference between our guiding manner and Dynamic Convolution \cite{chen2020dynamic}. Our method is significantly superior to Dynamic Convolution \cite{chen2020dynamic} because the average pooling operation in its attention mechanism leads to severe loss of spatiotemporal information. However, our mask mechanism and attention mechanism better leverage the characteristics of videos.

\begin{table}
\centering
  \caption{Ablation study on Frame Number on K400 and SSv2.}
  \label{tab:frame_number}
  \begin{tabular}{lcccc}
    \toprule
    \multirow{2}{*}[-0.3em]{ Frame } & \multicolumn{2}{c}{ K400 } & \multicolumn{2}{c}{ SSv2 } \\ \cmidrule(lr){2-3} \cmidrule(lr){4-5}%
     & Top-1 & Top-5 & Top-1 & Top-5 \\
    \midrule
    32 w. D & 78.6 & 93.5 & 68.2 & 90.9 \\
    32 w/o D & 79.5 & 94.2 & 68.9 & 91.2 \\
    64 w. D & 79.8 & 94.1 & \bf{69.6} & \bf{91.7} \\
    64 w/o D & \bf{80.2} & \bf{94.3} & 69.5 & 91.6 \\
    \bottomrule
  \end{tabular}
\end{table}
\begin{table}
\centering
  \caption{Ablation study on Convolution in Self-Attention on K400 and SSv2.}
  \label{tab:ab_msaconv}
  \begin{tabular}{lcccc}
    \toprule
    \multirow{2}{*}[-0.3em]{Conv in MSA} & \multicolumn{2}{c}{K400} & \multicolumn{2}{c}{SSv2} \\ \cmidrule(lr){2-3} \cmidrule(lr){4-5}%
     & Top-1 & Top-5 & Top-1 & Top-5 \\
    \midrule
    2D & \textbf{79.5} & \textbf{94.2} & \textbf{68.9} & \textbf{91.2} \\
    3D & 79.0 & 93.8 & 68.5 & 91.1 \\
    \bottomrule
  \end{tabular}
\end{table}

\textbf{Frame Number.} 
Based on the experiments in Non-local \cite{wang2018non} and our hypothesis, we investigated whether increasing the number of frames can improve performance, as shown in Table~\ref{tab:frame_number}. While performance does improve, overfitting is observed in SSv2 due to information redundancy between frames caused by the denser uniform sampling strategy.

\textbf{Convolution in Self-Attention.} 
As shown in Table~\ref{tab:ab_msaconv}, within the variants of self-attention, the use of 2D convolution is also more effective than 3D convolution. A possible reason for this could be that the inflating operation may disrupt the features learned by self-attention in pretraining. We hope that these findings can provide insights for future network designs.

\subsection{Visualization}
We provide more visualization details in this subsection.

\begin{figure*}[htbp]
  \centering
  \begin{subfigure}[t]{0.45\linewidth}
    \centering
    \includegraphics[width=\textwidth]{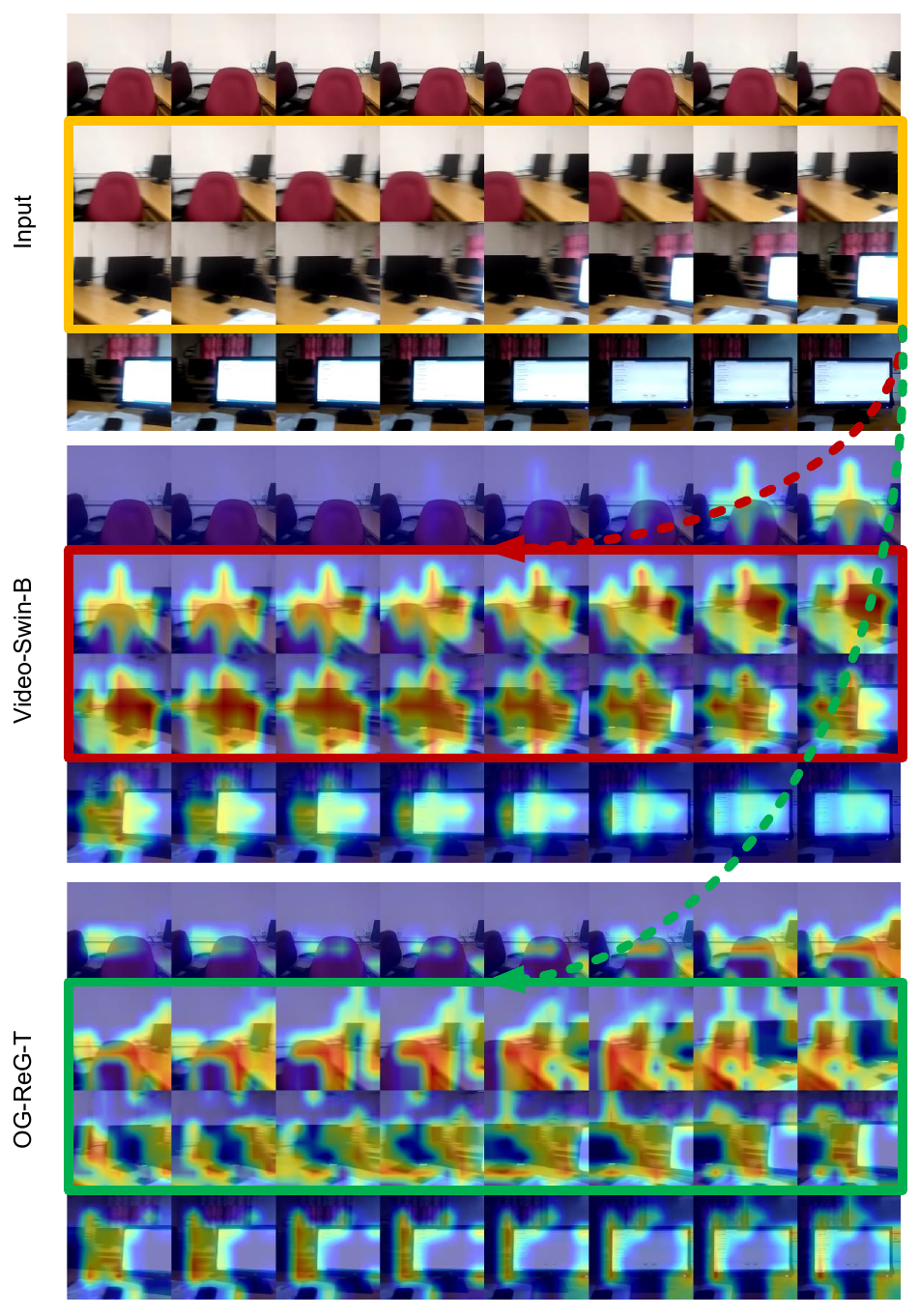}
    \caption{\textit{Turning the camera right while filming a chair.}}
    \label{fig:gradcam1}
  \end{subfigure}
  \hfill
  \begin{subfigure}[t]{0.45\linewidth}
    \centering
    \includegraphics[width=\textwidth]{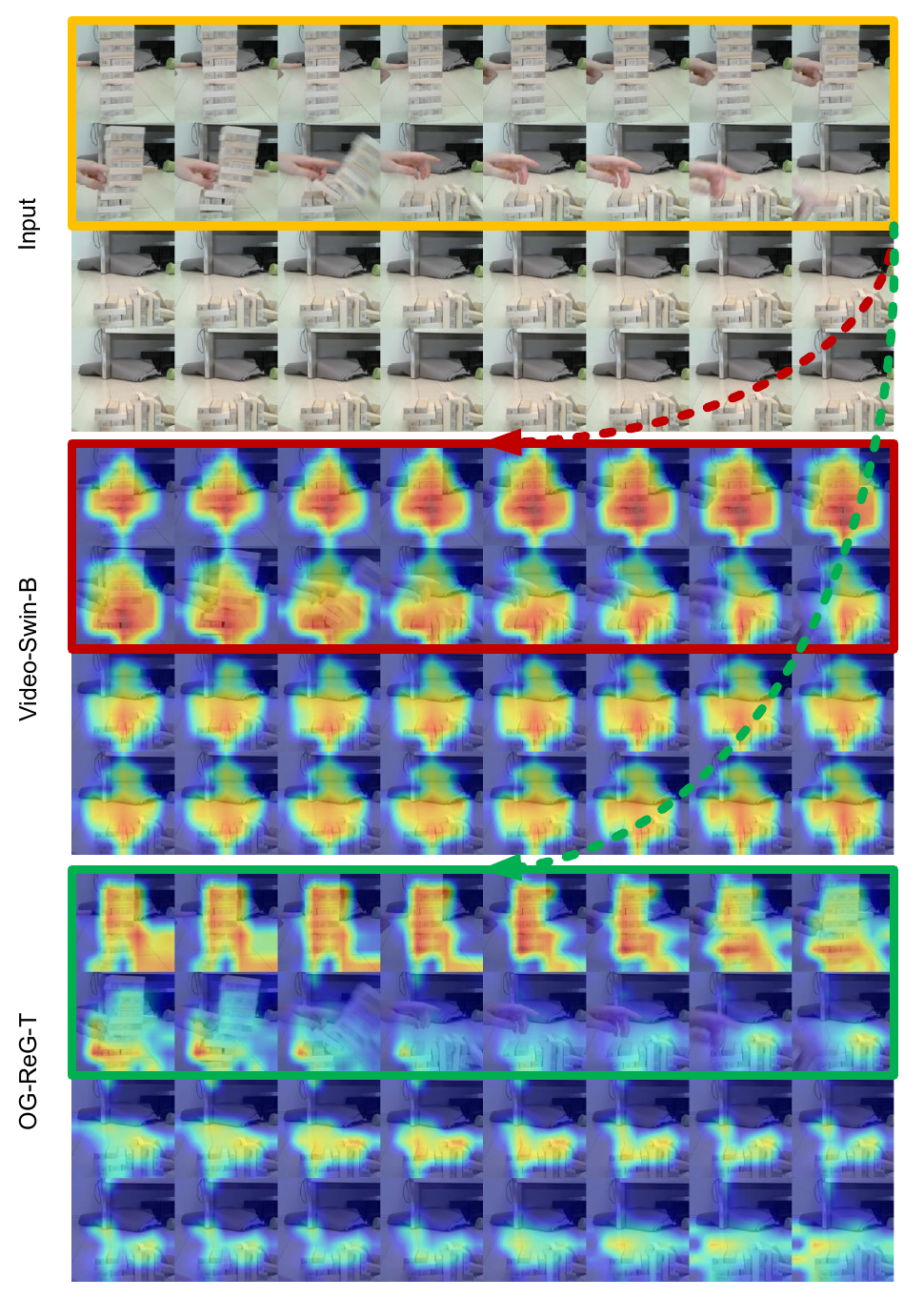}
    \caption{\textit{Poking a stack of blocks so the stack collapses.}}
    \label{fig:gradcam2}
  \end{subfigure}

  \vspace{10pt} 

  \begin{subfigure}[t]{0.45\linewidth}
    \centering
    \includegraphics[width=\textwidth]{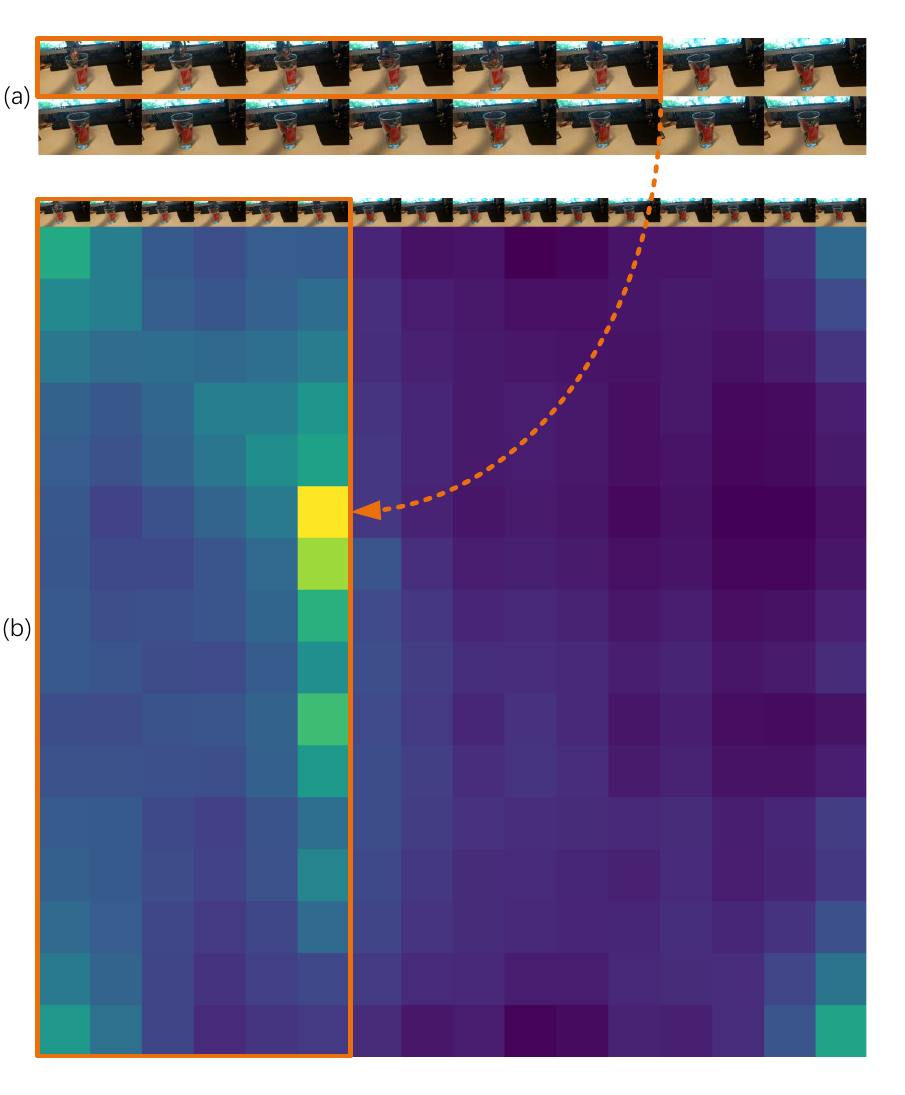}
    \caption{\textit{Dropping a keybound into a glass.}}
    \label{fig:tempo1}
  \end{subfigure}
  \hfill
  \begin{subfigure}[t]{0.45\linewidth}
    \centering
    \includegraphics[width=\textwidth]{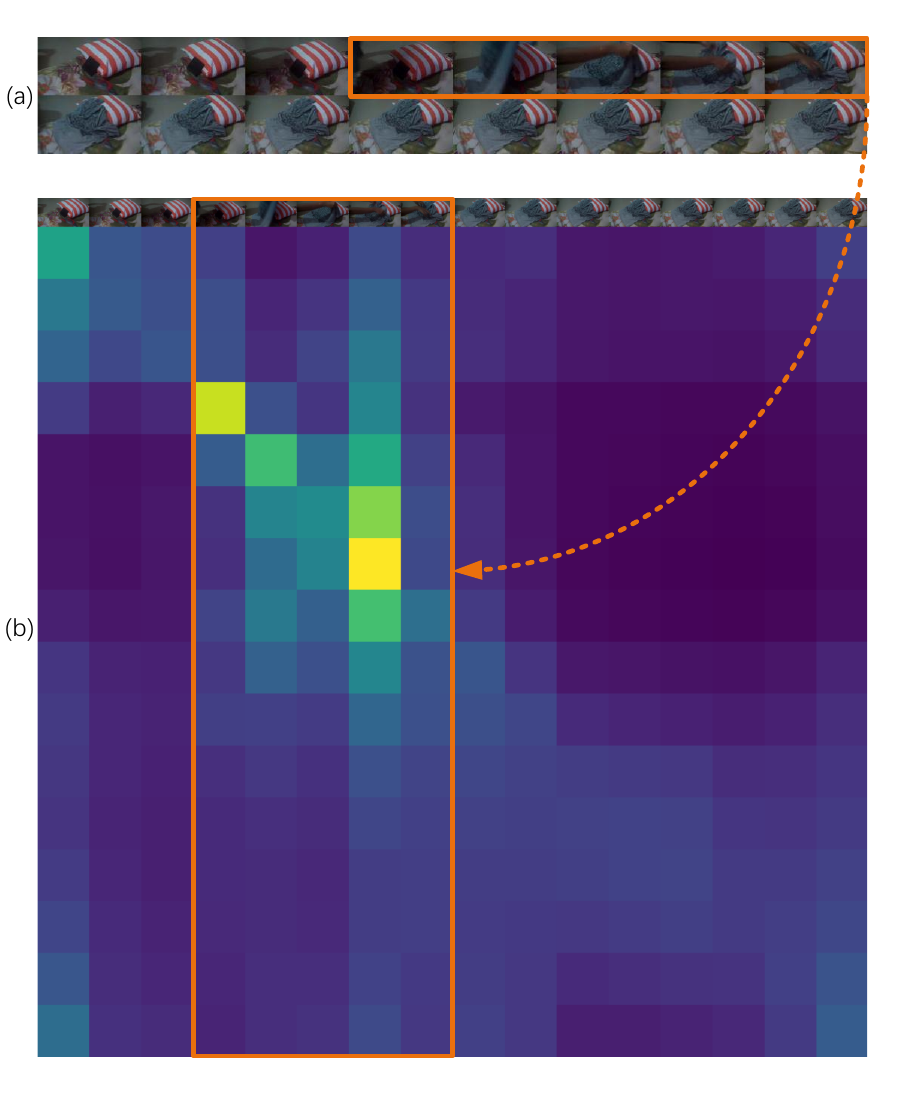}
    \caption{\textit{Covering a phone with a blanket.}}
    \label{fig:tempo2}
  \end{subfigure}

  \caption{Grad-Cam and Visual tempo Visualization.}
  \label{fig:four_subfigs}
\end{figure*}

\textbf{Grad-Cam.}
We select two representative scenarios to demonstrate the ability of our method to capture motion, with corresponding heatmaps generated using Grad-CAM \cite{Selvaraju_2017_ICCV} in the last layer. Figure~\ref{fig:gradcam1} illustrates a scene where the camera moves while the object remains stationary. The heatmap can be observed to follow the object's movement in the opposite direction of the camera, indicating our method's ability to perceive the camera's motion.
In Figure~\ref{fig:gradcam2}, the object is moving while the camera is stationary, and the heatmap changes with the object's collapse, indicating the ability of our method to perceive the object's motion. 
In contrast, the heatmap of Video-Swin \cite{liu2022video} shows a fixed pattern, focusing on the center of the image, and it does not change with the movement of objects and the camera in time and space and blindly compares
the similarity of all tokens in all.
It strongly supports our motivation.

\textbf{Visual tempo.}
We provide additional visualizations of the visual tempo to further demonstrate the effectiveness of the frame similarity. As shown in Figure~\ref{fig:tempo1}, our method can effectively detect the end boundaries of fast-tempo actions, exemplified by the moment the keybound is dropped into the glass at the sixth frame. Additionally, as depicted in Figure~\ref{fig:tempo2}, our method can accurately capture the duration range of actions with a longer temporal extent, for instance, such as the motion of reaching out and retracting the arm from the fourth to the eighth frame.

\section{Conclusions}\label{sec5}
In this paper, we analyze the importance of spatial and temporal information at different time scales, and propose a method with SoDA and NDConv, which is similar to the glance and gaze and can efficiently process coarse-grained overall spatiotemporal information on clip level while supplementing the local detailed information required for glance on frame level. In spite of these observations, open problems remain. The proposed is still not as efficient as the human visual system in predicting and attending to important spatiotemporal cues. We will prioritize the combination of time and space at different stages of the network based on some findings and efficiently model time and selectively process some important spatial information \cite{buch2022revisiting} as future work. Predicting important events' timing and location, as humans do, is crucial for future video models.

\bibliographystyle{cas-model2-names}
\bibliography{main}
\end{document}